\documentclass[preprint,12pt,number]{elsarticle}

\usepackage{amssymb}
\usepackage{amsmath}
\RequirePackage{graphicx}
\RequirePackage{latexsym,ifthen,rotating,calc,textcase,booktabs,color,endnotes}
\RequirePackage{amsfonts,amssymb,amsbsy,amsmath,amsthm}
\usepackage{mdframed}  %

\usepackage{adjustbox}

\usepackage[normalem]{ulem}
\newenvironment{acks}[1]%
{\subsection*{Acknowledgements}\noindent #1}%
{}

\usepackage{moreverb,url}

\usepackage[colorlinks,bookmarksopen,bookmarksnumbered,citecolor=blue,urlcolor=red]{hyperref}

\graphicspath{{figs/}}
\usepackage{xurl}
\usepackage{float} \usepackage{tcolorbox}
\tcbuselibrary{fitting}

\usepackage{colortbl}
\usepackage{tikz}
\usetikzlibrary{arrows.meta, positioning}
\usepackage{newpxmath}
\usepackage[normalem]{ulem}

\usepackage[utf8]{inputenc}   
\usepackage[T1]{fontenc}

\definecolor{comp}{HTML}{FFDDC1}
\definecolor{mix}{HTML}{C1FFC1}
\definecolor{found}{HTML}{C1D4FF}

\newtheorem{definition}{Definition}

\newtheorem{postulate}{Postulate}

\newcommand\BibTeX{{\rmfamily B\kern-.05em \textsc{i\kern-.025em b}\kern-.08em
T\kern-.1667em\lower.7ex\hbox{E}\kern-.125emX}}

\journal{INFFUS}

\begin{document}

\begin{frontmatter}

\title{Data Science: a Natural Ecosystem} %

\author[1,4]{Emilio Porcu}
\author[1]{Roy El Moukari}
\author[2]{Laurent Najman\corref{cor1}}
\author[3,4]{Francisco Herrera}
\author[4]{Horst Simon}

\affiliation[1]{department={Department of Mathematics}, organisation={Khalifa University}, city={Abu Dhabi}, country={UAE}}
\affiliation[2]{organisation={Univ Gustave Eiffel, CNRS, LIGM}, postcode={F-77454}, city={Marne-la-Vallée}, country={France}}
\affiliation[3]{
organisation={Dept. of Computer Science and Artificial Intelligence, 
DaSCI Andalusian Institute of Data Science and Computational Intelligence,
University of Granada}, city={Granada}, country={Spain}}
\affiliation[4]{organisation={ADIA Lab}, city={Abu Dhabi}, country={UAE}}

\cortext[cor1]{Corresponding author}

\begin{abstract}
This manuscript provides a {systemic and data-centric} view of what we term {\em essential} data science, as a {\em natural} ecosystem with challenges and missions stemming from the {fusion of} data universe with its multiple combinations of the $5$D complexities (data structure, domain, cardinality, causality, and ethics) with the phases of the data life cycle. Data agents perform tasks driven by specific {\em goals}.
The data scientist is an abstract entity that comes from the logical organization of data agents with their actions. Data scientists face challenges that are defined according to the {\em missions}. We define specific discipline-induced data science, which in turn allows for the definition of {\em pan}-data science, a natural ecosystem that integrates specific disciplines with the essential data science.  \\
We semantically split the essential data science into computational, and foundational.  \\
By formalizing this ecosystemic view, we contribute a general-purpose, fusion-oriented architecture for integrating heterogeneous knowledge, agents, and workflows—relevant to a wide range of disciplines and high-impact applications.
\end{abstract}

\begin{highlights}
\item {Fusion of the complexities of the data universe (five levels of complexity) with the data life cycle (five phases).}
\item {Holistic approaches to a natural ecosystem: the Essential Data Science.}
\item Systemic approach to integration of the Essential data science with other disciplines: the Pan Data Science. 
\item {Fusion of Mission $\&$ Visions with task-based approaches to define data agents and data scientists.}
\end{highlights}

\begin{keyword}

Data Agent \sep Data Universe  \sep Mission \sep Route-to-Discovery.
\end{keyword}

\end{frontmatter}

\section{Introduction}

Data Science sits at the intersection of statistical modeling, algorithmic prediction, and domain decision-making. Prior framings ({\em e.g.}, Breiman’s two cultures, Donoho’s “greater data science,” and the Fourth Paradigm) clarify pieces of the landscape, but leave open how missions translate into concrete challenges and tasks across heterogeneous data contexts. This paper proposes an “Essential Data Science” (EDS) ecosystem that operationalizes that translation via five complexity dimensions (5D) and five data-life-cycle phases (DLC). Hence, our paper integrates, rather than contradicts, previous contributions by \cite{breiman2001statistical, liberman2009effect, LectureA, hernan2019chance, hey2009fourth, donoho2024data}. In particular, our scope complements process-centric views \citep{aalst2016process} by focusing on  the chain {\em mission $\to $ challenge $\to$ task atomization} across domains, beyond event-log analytics.

\subsection{Between Convergence and Ecosystems}
Data science has traditionally been seen as a superset of statistics, machine learning and computer science, with the addition of {\em technological findings that would allow for scalability}. This was {\em motivated by commercial rather than intellectual developments} \citep{donoho201750}. 
 The dawn of data science as a discipline came with contributions from \citet{tukey1962future}, \citet{cleveland2001data} and \citet{breiman2001statistical}. The discussion is  mainly about
 blending statistics, machine learning, and computer science properly, while others had been fighting to claim either that {\em data science is statistics}\endnote{\url{ http://www.datascienceassn.org/code-of-conduct.html}}, or {\em data science is not statistics}\endnote{\url{https://statmodeling.stat.columbia.edu/2013/11/14/statistics-least-important-part-data-science/}}; 
 and boosting the development of data science through the so-called common task framework (CTF). By CTF, we mean benchmark-driven evaluation, with fixed training/validation splits, and a scoring leaderboard popularized in ML competitions and standardized tasks \citep{breiman2001statistical, donoho201750}; our EDS relates to, but is not limited by, CTF settings.
  \\
We claim that 
we run the risk of approaching a {\em winter era} in data science, where, despite multiple warnings that {\em data science is not big data}, the main focus has been on computation within the realm of huge (real or artificial) datasets, technological findings, and storytelling, with little attention to foundational aspects. 
 \citet{meng2019data} defines data science as an artificial ecosystem  that {\em is not a discipline by itself}, because of the large portfolio of skills that are required to do data science. As a consequence, 
 \citet{jordan2019artificial} and \citet{meng2019data} agree that artificial ecosystems are prone to disasters.

This paper argues that 
\begin{enumerate}
\item Convergence on a definition of data science and data scientist can be achieved by starting from 
what we call the {\em data universe} (a triplet that contains data, information and knowledge), inspiring missions and subsequent challenges.
\item The complexity of challenges requires their {\em atomization} into simpler 
goals and related tasks. This in turn allows us to define data agents as atomic entities with tasks and goals and, consequently, defined skills. Agents can be humans or machines.
\item Atomic agents are organized in a system that is consistent with challenges, missions, and complex interactions. This system becomes a new entity called {data scientist}. 
\item Data scientists deliver results in the form of data universe discovery, which consequently updates the data universe. 
\item Our data-centric view culminates with the claim that data science is not only a discipline per se, but a {\em natural ecosystem}, {deeply rooted in the complexities of the data universe and structured through a multidimensional fusion of agents, tasks, and disciplines.}
\end{enumerate}

This work defines the {\em essential} data science (EDS) as the {discipline} {that is} involved in the challenges of the universe of data. After identifying the pillars of the EDS, we define
specific {discipline}-induced data sciences through the integration of specific {discipline} ({\em e.g.}, medicine, finance, climate) with the EDS. 
The collection of all these possible disciplines, in concert with the interactions between them, is termed {\em pan}-data science. In particular, we use EDS for the core mechanism that maps missions to challenges and tasks through (5D, DLC) pairings. Pan-Data Science denotes the broader ecosystem, including other disciplines in concert with institutions, incentives, and governance.

\subsection{{Making a Difference}}
Are there similar papers in the data science landscape?
The following {elements highlight} our distinctive contributions. The paper:
\begin{enumerate}
\item provides a holistic view. Our starting point is the data universe, that triggers all subsequent missions.  The cultures of data science (see Section \ref{sec2}) start instead with a specific mission (see Section \ref{sec2}).  
\item is discipline-independent. It applies to any discipline interested in data science. And, basically, all disciplines are! 
\item is Artificial Intelligence (AI)-independent, in the sense that the paper does not rely on any particular definition of AI, nor any particular AI technique. 
This will help to alleviate the
confusion present in the literature.
\item has philosophy, ethics, and mathematics, as the {drivers} of its view. 
\end{enumerate}

\subsection{Take Away {for the Information Fusion Community} }
{Broadly speaking, an ecosystem is {\em a complex network or interconnected system}\endnote{Available at \url{https://languages.oup.com/google-dictionary-en/}.}. While this definition fits quite reasonably to what we are proposing, the additional argument we defend is that of {\em natural}, intended here as {\em existing in nature} -- and we add: not entirely caused by humans, or machines. We defend this assertion in view of the following findings. }  
\begin{enumerate}
\item A holistic view of the data-centric universe allows defining missions, challenges, and their subsequent atomization into tasks and goals.
\item Atomization of tasks and goals  mirrors our proposal of systemic organization of data universe into (a) complexities (we prioritize $5$ of them) and (b) phases (called vertices) of the data life cycle (and we find $5$ of them).
\item Data agents are atomic entities that might be humans or machines. Data scientists are logically organized entities of data agents.  
\item Data agents have tasks and goals, data scientists have challenges coming from {\em missions} inspired by the data universe. 
\item The EDS is integrated with specific disciplines to build discipline-induced data sciences. The {\em pan}-data science is the collection that includes each of these discipline-induced disciplines in concert with their logical arrangements. 
\end{enumerate}

{In conclusion, we provide  two major innovations that are relevant to the Information Fusion community: \\
1. A {\em formal model} of fusion in data science, where the  five dimensions of complexity interact with the five-phase of the data life cycle.\\
2. An agent-based architecture for the integration of human with algorithmic agents into a fusion system that is coherent and well driven, and that adapt to missions that are inherently domain-specific.

{T}he paper is organized as follows: Section \ref{sec2} explores the dominant cultures in Data Science. Section \ref{sec3} delves into the data universe, {semantically organized into combinations of  five dimensions with five phases of what we term data life cycle}. Section \ref{sec4} focuses on the mission, vision, tasks, and goals, which are then used in Section \ref{sec5} to define the EDS, data agents, and data scientists. Finally, we conclude with a brief discussion and key takeaways from the paper.

\section{The {\em Dominant} Cultures in Data Science} \label{sec2}
This section is largely expository and provides an overview of the most influential cultures within more than 60 years of data science. Historically, each culture coalesced around prevailing missions ({\em e.g.}, scientific explanation {\em vs} large-scale prediction). We use this lens descriptively to explain observed transitions rather than to impose a normative taxonomy.
\begin{enumerate}
    \item {\em Breimann's Culture}: in his seminal paper, \citet{breiman2001statistical} establishes the dichotomy {\em generative} versus {\em predictive} modeling. The former is the heritage of the classical statistical paradigm, oriented to the construction of stochastic models, to then make inference about the underlying mechanism generating the data. For this, certain theoretical assumptions on the underlying model are needed. Predictive modeling has been widely adopted by the empirical machine learning community. \citet{donoho201750} argues that the {\em secret sauce} boosting predictive approaches is the celebrated {\em common task framework}, that has been especially successful among machine learners, and mostly ignored by the orthodox statisticians. \citet{liberman2009effect} defend the common task framework on the basis of the appreciable improvements that are coming through a collection of small increments. 
\item {\em Donoho's Transition} is based on the path: from {\em math First} into {\em data First} \citep{LectureA}. The former being boosted during the seventeenth century, and the latter being {specific to the} 1980ies. The purported change of paradigm entails 
 (a) Models versus Problems; (b) Derivations versus Scaling; (c) Analysis versus Measurement; (d)  Understanding versus Performance, and (e) Discourse versus Products.
\item {The {Hernán-Hsu-Healy} approach (HHH throughout, see \citealp{hernan2019chance}) to data science is centered on causal inference. 
The potential of data science is unexplored within the framework of observational data. The authors' transition is causality-centered, and the
driver is the ability to create counterfactuals. While rating data science as an {\em odd combination of words}, {\em HHH} theory implicitly defines data science through a prioritization of three main tasks, being 
 description, prediction, and counterfactual prediction. While automated predictive models can yield substantial insights, causal inference uniquely relies on expert knowledge and requires more than just algorithms and data. The expert's insight guides the analysis, informs the creation of credible assumptions, and provides the necessary framework to interpret complex relationships. Causal inference lacks a universally accurate algorithmic solution. Hence, human expertise is essential for establishing the credibility and validity of causal conclusions. The {\em HHH} view underscores the importance of integrating domain knowledge with data science methodologies to effectively advance our understanding of causality in complex systems.}
\item {Jim Gray} \citep{hey2009fourth}  terms {\em Fourth Paradigm} the switch to data-intensive-science methods, integrating empirical, theoretical and computational science with the direct goal of leveraging massive datasets and advanced computational tools to generate insights. Unlike previous paradigms, it prioritizes the following three key components: (a) data capture; (b) data curation; and (c) data visualization, which are to be combined with supporting infrastructures such as data repositories, analytical tools, and interoperable publications. This system will allow
increasing information speed, and will support reproducibility. 
\item Donoho's {\em hidden superpower} in computational data science \citep{donoho2024data} is triggered by a {\em singularity}. This is in turn favored by 
 {\em frictionless reproducibility} (FR).
{Empirical} machine learning is the leading adherent field and the hidden superpower is
adherence to FR practices; in turn, these practices are responsible for the
striking and surprising progress in AI that we see everywhere; they can be learned and
adhered to by researchers in whatever research discipline, automatically increasing the rate of research discoveries. 
\end{enumerate}

\smallskip
\noindent
Clearly, the five cultures are not incompatible and have notable intersections. \\
We note that all these cultures are inspired by a {\em mission}. The mission is a {\em transition}
\begin{enumerate}
\item from generative into predictive paradigm \citep{breiman2001statistical};
\item from {\em math first} into {\em data first} \citep{LectureA}; 
\item from prediction into counterfactual prediction \citep{hernan2019chance};
\item from a classic to a new scientific paradigm focused on data-intensive-science methods \citep{hey2009fourth};
\item from {\em frictional} into {\em frictionless} reproducibility to become {\em as fast as the speed of light} in data universe discoveries \citep{donoho2024data}.
\end{enumerate}

\section{Data Universe, the  5Ds, and the  Data Life Cycle} \label{sec3}
 {The distinction between data, information, and knowledge has been extensively debated in the literature. While the discussion remains unresolved, it lies beyond the scope of this paper.}
 Here, we shall adopt the classification based on decision-making, as in \citet{aamodt1995different}, for which:
\begin{enumerate}
    \item Data are syntactic entities, {\em i.e.}, the input to the initial step of decision-making; 
    \item Information is {\em data with meaning};
    \item Knowledge is learned information. 
\end{enumerate}
In turn, \citet{aamodt1995different} elaborate on knowledge in terms of roles,  specifically (a) interpretation, that is, transforming data into information, (b) elaboration, which allows inferring new information from already existing one, and (c) learning, through which new knowledge is achieved. According to \citet{leonelli2019data}, “there is no such thing as raw data.”
\begin{definition}
    \label{5D} The data universe is a triplet that contains data, information, and knowledge.  
\end{definition}
The data universe is a dynamic entity, as it is constantly {\em updated}. This concept will become apparent during subsequent sections. \\
We propose a semantic organization of the data universe according to:

\begin{enumerate}
\item[(a)]  five {\em dimensions} of {data} {\em complexities}, denoted {\bf $5$D} throughout, and 
\item[(b)] five phases of the {\em data life cycle}, denoted {\bf DLC}. 
\end{enumerate}
 We remark that our five DLC phases are decision-anchored, and complementary to prior life-cycle framings \citep{berman2018realizing}.

When adding a subscript $i$ (a natural number), we term { $5$D$_i$ the $i^\text{th}$ dimension of 5D and DLC$_i$ the phase $i^\text{th}$ of DLC}. \\
We provide here a succinct description of the $5$Ds.

\begin{enumerate}
\item {\em Data domain} (5D$_1$) defines the context, origin, and intended application of the dataset being analyzed. It encapsulates the specific field or subject area from which the data is sourced and where it holds relevance. For instance, a dataset in the healthcare domain might contain patient records, medical images, or genomic sequences, while one in the finance domain could include stock prices, transaction histories, or economic indicators. Understanding the data domain is crucial for framing the problem, identifying meaningful features, and ensuring that the analysis aligns with domain-specific goals. Furthermore, it often influences the types of questions that can be answered, the metrics for evaluation, and the ethical considerations that must be addressed.

\item Data structure (5D$_2$) refers to the way  data are organized and represented, which directly impacts processing and analysis. The advent of very complex  dataset  brought to us the concept of a datum, as a potentially abstract entity that can move over any function space. This concept is beautifully condensed by \citet{leonelli2019data} in a philosophical framework. This happens, for instance, when the data is  tabular data with rows and columns, time-series data with temporal dependencies, unstructured text, images, or graphs, etc. Data structure also considers dimensionality, sparsity, and scalability, which are important factors in selecting algorithms and designing computational workflows.  
\item Data cardinality (5D$_3$)  indexes how much and how heterogeneous the data are. Understanding data cardinality is critical in data science as it impacts storage, processing, and modeling strategies.
as fitness-for-use (completeness, accuracy, bias, timeliness). 
\item Data causality (5D$_4$) {refers to the study and identification of cause-and-effect relationships within data. It goes beyond statistical correlations to determine whether one variable directly influences another. Understanding causality is critical for making informed decisions, as it allows for interventions that can change outcomes, rather than simply observing patterns.} The data universe is an extremely complex and dynamical system. Humans are designed to have perceptions about relationships, that are often depicted in terms of associations, correlations, and dependencies. These concepts have been well understood by the probability $\&$ statistics community since more than a century. {\em Causality}, despite being a very natural concept, is relatively new in terms of mathematical formalism. The reader is referred to the books by \citet{pearl2009causality,Hernan-Robins-2020} and \citet{de2023causal} for an extensive account. Causality is oriented to deducing the effect of hypothetical interventions or changes in a system, which is quite different from observing associations in the data. \citet{hernan2019chance} argue that data science entails creating {counterfactual predictions}, answers to {\em what if} scenarios, by leveraging observational or experimental data to establish causal relationships rather than just association. In contrast with prediction tasks, which rely on the mapping of inputs to outputs, 5D$_4$ demands domain expertise to accurately model and interpret the causal layout of the data, typically using unverifiable assumptions and structural knowledge about the system studied.\newline
The key components of 5D$_4$ are
\begin{enumerate}
    \item[a.] Counterfactual predictions: {\em what if the world was different?} ({\em e.g.}, what if a {medical} treatment were given or not). Causal Inference aims to estimate what the outcome would have been under alternative conditions.
    \item[b.] Structural knowledge and expert input: unlike prediction, causal inference cannot rely only on data. Structural knowledge and expert input are essential to correctly adjust for confounding variables. \citet{hernan2019chance} use the example of maternal smoking and infant mortality, where factors like the “birth-weight paradox” should be taken into consideration.
    \item[c.] Importance of {\em confounding adjustment}: causal inference must identify and control confounding variables influencing both the treatment and the outcome. For example, in healthcare, lifestyle factors affecting both smoking during pregnancy and infant health outcomes must be addressed to isolate the causal effect of smoking~\citep{hernan2019chance}.
    \item[d.] Randomization and observational Data: while randomized control trials (RCT) provide unbiased causal estimates, they are not always feasible. 5D$_4$ in observational studies relies on assumptions and statistical adjustments to approximate what a RCT might reveal. Techniques such as the g-formula and inverse probability weighting are applied to achieve this, as shown in~\citet{hernan2019chance} for healthcare observational data.
\end{enumerate}
DLC$_4$
  aligns analyses with mission KPIs (e.g., AUROC/recall for alerts, $\Delta$-CRPS for probabilistic forecasts, equity gaps across groups). Analyses not tied to KPIs are flagged as exploratory.
\item Data ethics (5D$_5$) is the branch of ethics that addresses the responsible collection, storage, processing, analysis, sharing, and use of data. It emphasizes the principles and values that ensure data practices are fair, transparent, and respectful of individuals’ rights and societal well-being.  It is a rapidly evolving discipline \citep{saltz2019data,barocas2017engaging,okorie2024ethical} that studies data, algorithms, and practices to support the so-called {\em morally good} decisions. \citet{floridi2016data} argue that ethical inquiries should no longer be info-centric but rather data-centric, so to enlarge the spectrum of ethical inquiries to include even non-structured data. This opens up a wealth of moral dimensions for all kinds of data.\\
Several crucial aspects are to be taken into account in order to guarantee ethical success within the data science framework \citep{saltz2019data,barocas2017engaging,okorie2024ethical}. Some of them are social acceptability and legal prohibition. As it stands now, data ethics is more of a philosophical discipline and, of course, a subset of pure ethics. The shift from information ethics to data ethics is probably more semantic than conceptual. Yet, it serves very well the scopes of this paper, as the data universe is being organized according to data, information, and knowledge, and we argue that different levels (and challenges) of ethics are actually related to those. 
\end{enumerate}
Table \ref{tabcard} illustrates the main differences between data quality and data cardinality.

\begin{table}[t] 
\centering
\caption{\label{tabcard}Cardinality vs.\ Quality in the 5D/DLC scheme. Cardinality concerns scope/diversity/scale of data; Quality concerns fitness-for-use (accuracy, completeness, bias, timeliness).}
\begin{tabular}{p{0.47\linewidth} p{0.47\linewidth}}
\hline 
\textbf{Cardinality } & \textbf{Quality} \\
\hline
\emph{Measures:} How much and how heterogeneous the data are
(spatial/temporal coverage, sample size $n$, duration $T$, number of modalities/sources).
&
\emph{Measures:} Whether data are fit for the mission
(completeness, accuracy/calibration, bias/fairness, timeliness, drift/stability, correct linkage/geocoding). \\[0.8em]
& \\
& \\
\emph{Questions/Actions:} Do we have enough sensors/years/modalities to answer the KPI? Should we widen coverage or add a new source?
&
\emph{Questions/Actions:} Are measurements reliable and comparable? Do we need QC, debiasing, calibration, imputation, or re-geocoding to meet KPI thresholds?  \\
\hline
\end{tabular}
\end{table}

\medskip
We now define the {\bf data life cycle} as the collection that includes the phases that feed the data universe. These are classified into the following:
\begin{enumerate}
    \item {\bf Capture} (DLC$_1$): this involves data acquisition, data entry, signal detection, and data extraction, to mention a few. 
    \item {\bf Categorization  } (DLC$_2$): %
    {Categorizing tasks to identify data for modeling and summarization. This  approach ensures that each task is analyzed in detail, considering factors such as complexity, dependencies, and expected outcomes, which ultimately enhances the precision and effectiveness of data-driven decision-making processes.}
    \item {{\bf  Quality}}  
    (DLC$_3$): {Good data quality is the bedrock of successful data science, ensuring accurate analysis, reliable predictions, and meaningful insights \citep{wang2024overview}. This falls under the well-known warehouse, cleaning, staging, data architecture, and data preprocessing \citep{garcia2015data}.}
    \item {\bf Analysis} (DLC$_4$): this entails exploratory data analysis, regression, prediction, text mining, and counterfactual prediction.
    \item {\bf Communication} (DLC$_5$): reporting, visualization, visuanimation, business intelligence, and decision-making. 
\end{enumerate}

Table \ref{tab:DLC_Dimensions} 
provides a rough classification of the pairs ($5D$.DLC) according to the pairing being mostly computational, foundational, or a blending of these two.

We are aware that this classification is not the only one possible and that much discussion has been opened around the subject. \citet{borgman2019lives} illustrates the {\em afterlife} of data, and data are often recycled to meet more challenges and new discoveries. A deeper data life cycle path is taken in~\citet{wing2019data}. \citet{borgman2017big} has a wonderful essay and a thoughtful analysis of large, small, and 'no' data. Her provocative questions would deserve an article per se: here, we recall that, according to her view, data are neither publications nor natural objects. Instead, data are representations. Finally, data sharing and reuse depend on knowledge infrastructures. We refer to her book for a profound discussion about data and scholarship.
\begin{table}[tb] 
\caption{\label{tab:DLC_Dimensions}A rough classification of the data universe.}
\centering
\begin{tabular}{|>{\centering\arraybackslash}p{2.5cm}|>{\centering\arraybackslash}p{1.5cm}|>{\centering\arraybackslash}p{1.5cm}|>{\centering\arraybackslash}p{1.5cm}|>{\centering\arraybackslash}p{1.5cm}|>{\centering\arraybackslash}p{1.5cm}|}
\hline
 & DLC$_1$  & DLC$_2$  & DLC$_3$  & DLC$_4$  & DLC$_5$ \\
\hline
Data Structure & \cellcolor{comp} & \cellcolor{mix} & \cellcolor{comp} & \cellcolor{mix} & \cellcolor{comp} \\
\hline
Data Cardinality & \cellcolor{comp} & \cellcolor{comp} & \cellcolor{comp} & \cellcolor{comp} & \cellcolor{comp} \\
\hline
Data Domain & \cellcolor{mix} & \cellcolor{mix} & \cellcolor{mix} & \cellcolor{mix} & \cellcolor{mix} \\
\hline
Data Causality & \cellcolor{mix} & \cellcolor{found} & \cellcolor{mix} & \cellcolor{found} & \cellcolor{found} \\
\hline
Data Ethics & \cellcolor{found} & \cellcolor{mix} & \cellcolor{found} & \cellcolor{found} & \cellcolor{found} \\
\hline
\end{tabular}

\vspace{0.6em}
\begin{tabular}{l l l l l l}
\cellcolor{comp} \hspace{1em} & Computational &
\cellcolor{found} \hspace{1em} & Foundational &
\cellcolor{mix} \hspace{1em} & Blending 
\end{tabular}
\label{table1} %

\end{table}

We conclude this section with an example that illustrates those definitions on a concrete case.

\begin{mdframed}[linecolor=black!15,backgroundcolor=black!03,roundcorner=6pt]
\noindent\textbf{Running Example — Urban Heat \& Health.} \\
For this worked example, the mission is to 
{quantify and reduce heat–health risk in a given  City (call it $\mathcal{C}$) during summer extremes.
The Data Universe includes near-surface meteorology (air temperature, humidity, wind), built/green infrastructure,
remotely-sensed land–surface temperature, fixed/mobile sensors, and de-identified emergency visits.}

\medskip
\noindent\textit{5D: Complexity Dimensions --}
Domain ($5D_1$) includes the  City $\mathcal{C}$; stakeholders: health authority, city planning. The primary KPI is calibrated exceedance risk for heat-stress and heat-related ER visits. \\
The structure (5D$_2$) is represented by an 
  irregular urban geometry: in fact, we have the typical structure of a metric graph, with coast/basin boundaries. Data are multi source
 (reanalysis grid, satellite pixels, point sensors) on misaligned supports. Regarding the cardinality (5D$_3$), data might be
  hourly fields over $10$–$100$\,m grid (millions of cells over a season), $\sim$dozens of sensors, multi-year archives, and heterogeneous modalities (satellite, station, administrative records). \\
  Causality ($5D_4$) might also play part of the game whenever provide models that allow to identify a causal action of hurban heat on health (typically, graphical causal models are used). Finally, ethics ($5D_5$) is part of the story in terms of
  privacy-preserving health aggregates (HIPAA/GDPR-compliant), fairness across districts, transparency of thresholds, reproducible pipelines, and documented uncertainty. \\
Our worked examples continues with the identification of DLC components. As for {\tt DLC}$_1$, specifying KPIs, space-time resolution, as well as choosing physics-aware downscaling ad Bayesian linkage are part of it. In terms of capture,
  typically ERA5/ERA5-Land data are collected. Data might come from station networks, mobile sensors, Landsat/ECOSTRESS LST, building/greenery GIS layers, and de-identified ER visits (temporal aggregation, ICD codes). \\ Regarding Data quality, this needs a careful check through
sensor drift checks, gap-filling, satellite debiasing, geocoding audits, and missingness patterns, in concert with maintaining a \emph{data sheet} with lineage and limits. As for analysis, this includes a wealth of alternatives that go from 
  downscaling meteorology via conservative advection–diffusion on urban meshes to creating exposure ensembles. Also, fitting Bayesian distributed-lag models linking exposure to ER visits, and validating with reliability/coverage and tail-focused scores. Finally, communication entails
  publishing ward-level risk maps and calibrated alerts with uncertainty bands, as well as document decision thresholds, equity diagnostics, and change-logs.
\end{mdframed}

\section{Mission, Vision, {Tasks and} Goals} \label{sec4}
There is a never ending philosophical debate regarding vision, purpose, paradigm, and mission, and most of the discussions are prompted by semantic roots. According to \citet{murphy1998concepts}, a mission is connected to the identification of a purpose that is deeper, that will make stakeholders proud to contribute. A simple example might be that of a company that {sells} healthy food: {\em we want to improve the quality of life of millions of people}. A mission is related to ideals, while vision is related to picturing the future. \\
 “Vision articulates a credible, realistic, attractive future for the organization”~\citep{bennis1985strategies}. 
A vision is based on a combination of cultures, beliefs, value systems, and  “values and beliefs are the most fundamental of the three elements of vision” \citep{quigley1994vision}. \\
According to \citet{murphy1998concepts}, there are three schools related to the definition of mission: (a) the strategic, (b) the philosophical, and (c) the military schools. Data science might be related to any of them. Certainly, philosophy plays a big role through ethics, because mission becomes a {\em cultural glue} \citep{murphy1998concepts}, that allows an organization to have an identity. Hence, mission encompasses cultural values, beliefs, and even emotions. \\
To put things into the context of data science, we state that the data universe is a strong candidate for inspiring missions. People become aware of problems through information, wherever it comes from, and we live in a hyper-connected world where missions can be inspired by anything inside the data universe. \\
We argue that missions trigger challenges, and these need to be strategically rationalized in order to map each challenge into a collection of tasks and goals. We call {\em atomization} the process that creates a hierarchical tree starting with the mission, creating the related challenges, and then the
tasks with respective goals related to each challenge. We assert that
\begin{postulate} \label{post1}
For each challenge, tasks and goals  %
are assigned according to the unordered pairs built as combinations of $5$D$_i$ complexities with any of the DLC$_i$, $i=1,\ldots,5$. 
\end{postulate}

The next assertion, shown in Figure \ref{data-univ}, prepares the subsequent section. 

\begin{figure}[H]
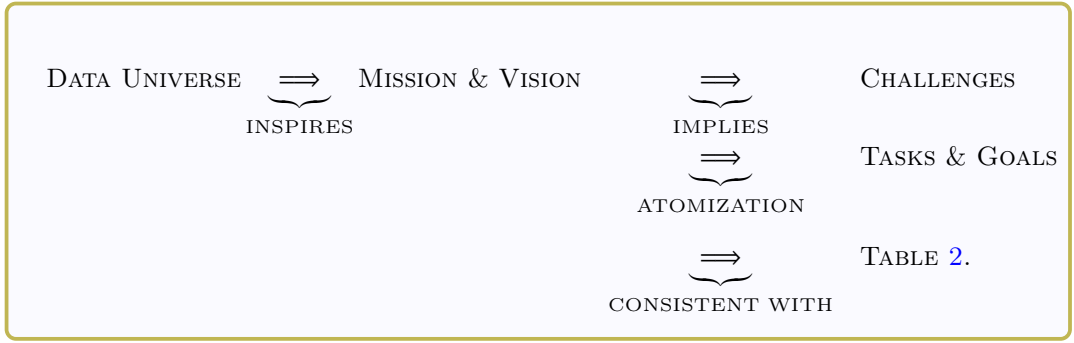

\begin{tcolorbox}[colback=blue!2!white,colframe=yellow!70!black,width=1.03\textwidth]
\footnotesize
\begin{eqnarray*}
 \text{{\sc Data Universe}} \underbrace{\Longrightarrow}_{\text{{\sc INSPIRES}}} \text{{\sc Mission $\&$ Vision}} &\underbrace{\Longrightarrow}_{\text{{\sc IMPLIES}}}& \text{{\sc Challenges}}  \\ & \underbrace{\Longrightarrow}_{\text{{\sc ATOMIZATION}}}& \text{{\sc Tasks $\&$ Goals}} \\
 \\ & \underbrace{\Longrightarrow}_{\text{{\sc CONSISTENT WITH}}}& \text{{\sc Table \ref{table1}. }}\end{eqnarray*}

 \end{tcolorbox}
\caption{Illustration of Postulate \ref{post1}, {connecting mission \& vision with taks \& goals}.} \label{data-univ}
\end{figure}

Data science and its protagonists are now introduced according to the above scheme. We aim to remark that missions may be defined top-down by stakeholders (policy/clinical need) or bottom-up from salient structures in the Data Universe. Our route-to-discovery is bidirectional: data can refine missions; missions can focus data capture efforts.

The purpose of atomization is to turn
 a mission into executable, testable tasks with owners, inputs/outputs, KPIs, and acceptance tests. We disentangle atomization through six steps. 
 \begin{enumerate}
     \item The first step is to state the mission and the KPIs, that is, define the decision you need to support and how success is measured (proper scores, coverage, equity).
     \item Select the 5D$\times$DLC pairings and identify which \emph{complexity dimensions} and \emph{life-cycle phases} are most consequential for the mission.
     \item Enumerate challenges from the chosen pairings, with a list of complete obstacles.
     \item For each challenge, define tasks with inputs, outputs, tools, and a responsible \emph{DS agent} (human or algorithmic).
\item Attach quantitative targets (e.g., CRPS, coverage, lead time, equity gap) and stop-criteria to each task.
\item Check dependencies, risks, \& critical path. 
Order tasks, log risks/mi\-ti\-ga\-tions, and define re-plan triggers (e.g., data access failure, drift). 
 \end{enumerate}
 We now provide a working example along the lines of the previous one.
\bigskip
\begin{mdframed}[linecolor=black!15,backgroundcolor=black!03,roundcorner=6pt]
\label{box:urbanheat}
{
\textbf{Mini-example: Urban heat alerts in City $\mathcal{C}$.}
Our mission is to issue neighborhood-level heat alerts that are both calibrated and fair, with uncertainty communicated transparently. Concretely, we target probabilistic forecast skill measured by CRPS and by quantile loss at $\tau\in\{0.90,0.95\}$; we require reliability close to nominal, with $90\%$ coverage within $\pm3\%$ both city-wide and by district; we insist on a minimum lead time of $24$ hours; and we track equity by capping the gap between false alarms and under-warnings at $5$ percentage points across districts. Within the 5D$\times$DLC framing, \emph{Structure} paired with \emph{Analysis} guides a physics-aware downscaling of coarse reanalyses (ERA5/ERA5-Land) using urban geometry, greenery and street-canyon GIS to produce hourly WBGT at $10$–$100$\,m resolution; \emph{Cardinality} paired with \emph{Capture} motivates the integration and quality control of heterogeneous streams (in situ sensors, satellite LST, emergency-room visits) into an aligned panel; \emph{Causality} paired with \emph{Analysis} enters through a Bayesian distributed-lag specification to link exposure ensembles with health responses; and \emph{Ethics} paired with \emph{Communication} shapes transparency (calibration diagnostics, versioned artefacts) and fairness auditing.

The work unfolds as a sequence of challenges turned into verifiable tasks. First, exposure downscaling over irregular urban morphology is addressed by constructing a modest ensemble that couples physical constraints with statistical learning so that PIT histograms remain stable and CRPS improves by at least $15\%$ relative to a baseline, while station-level RMSE drops by at least $20\%$. Next, heterogeneous data are harmonized: sensors and satellite products are reconciled temporally and spatially, missingness is held below $5\%$ after imputation, and each dataset is accompanied by a complete datasheet; only then do we fit the distributed-lag model, whose posterior predictive intervals must achieve the $90\%$ nominal rate within $\pm3\%$ overall and by district, with sharper upper-tail quantiles than a proxy-only baseline. With calibrated predictive distributions in hand, district thresholds for alerting are optimized under an explicit recall–precision trade-off (recall $\ge0.80$ at precision $\ge0.60$ for exceedance events) and are audited to keep the equity gap within the $5$\,pp limit while preserving a lead time of at least $24$\,h. Finally, risk maps, uncertainty bands and decision memos are published through a reproducible pipeline that refreshes products within $30$ minutes of new inputs and maintains a versioned change-log for traceability.

Dependencies are linear and transparent—data alignment precedes downscaling, which precedes health linkage, which precedes thresholding, which in turn precedes publication—so failure modes are easy to localize. Anticipated risks are managed in place: data-access constraints are mitigated by working with privacy-preserving aggregates; distributional drift is monitored by rolling reliability and quantile-loss checks; and computational load is contained via polynomial spectral filters and preconditioned solvers that keep turnaround consistent with the $24$-hour lead-time requirement. In sum, the example demonstrates how the 5D$\times$DLC pairings translate a civic mission into a verifiable workflow where skill, calibration, lead time and equity are treated as first-class, testable obligations rather than afterthoughts.
}
\end{mdframed}

\section{{Essential} Data Science, Data {Science} Agents, and Data Scientists} \label{sec5}

This section delves into the essentials of data science, exploring its foundational principles. We will also introduce data agents, entities that perform data-driven tasks, and data scientists, the skilled professionals who harness data agents.

\subsection{Essential Data Science}
We start by a rather general definition.
\begin{definition}
\label{def:DS} Essential Data Science (EDS) is the discipline that deals with the challenges triggered by missions and visions inspired by the data universe. 
\end{definition}
We sharpen the content of this definition as follows. 
\begin{postulate} \label{postA}
The data universe triggers missions. These inspire challenges within the EDS. In turn, the challenges inherent to EDS are {\em atomized} into tasks and goals  %
according to the unordered pairs, as in Table \ref{table1}. 
\end{postulate}
A semantic distinction is necessary to open up a further discussion. We virtually separate the EDS into foundational and computational. There is no neat separation (yet?) between them, but the separation is illustrative for several reasons, as detailed throughout.  
\begin{definition} \label{pillars} The  pillars  of EDS are statistics \& mathematics, empirical machine learning, and ethics \& philosophy. 
\end{definition}

Some readers may be surprised that ethics \& philosophy are proposed here as pillars of EDS. There are specific reasons for that: the emergence of data ethics as a foundational discipline inside data science requires integration of mechanistic views of the data planet with ethical and philosophical aspects~\citep{okorie2024ethical}. Another reason comes from the apparent transition from data-driven to data-centric, as advocated by \citet{leonelli2019data}. 

We can now combine Definitions \ref{def:DS} and \ref{pillars} to claim that 
the EDS functions, through a proper blending of the pillars, to solve the challenges coming from the data universe. 

In {computational} EDS,  the main pillar is  {\em empirical} machine learning, and the others have a minor impact. Computational data science works on the basis of (a) data universe sharing (b) code sharing, and code reproducibility, and (c) adopting challenge problems as a new paradigm powering the scientific research. The last being substantially the common task framework~\citep{breiman2001statistical, donoho2024data}.

\smallskip
The following definition opens up for further thought. 
\begin{definition}
\label{def:f-based-DS} Define ${\cal F}$ as a general notation for a discipline.  Then, we define ${\cal F}$-induced data science as the integration of the EDS within a specific ecosystem induced by ${\cal F}$.
\end{definition}

\begin{definition}
The Pan-Data science is the collection of (a) the essential data science, and (b) all the possible collections of ${\cal F}$-induced data sciences. 
\end{definition}

\subsection{Data Science Agents}
Who does data science? We start with a simple definition. 
\begin{definition}[DS Agent] \label{ds-agent}
{For a given challenge coming from a mission as per Postulate \ref{postA}, and for every goal coming from the corresponding atomization, the data science agent (DS agent or simply data agent) is an entity that performs actions to achieve the assigned task.}
\end{definition} 
There is a subtle duality in Definition \ref{ds-agent}, due to the fact that agents are determined according to their goals, which, in turn, come from challenge atomization. Apparently, an agent is needed to perform atomization. Hence, atomization can be both seen as an action (atomize to achieve goals from challenges) and as a goal (atomization is the goal to be achieved by a specific agent.) 

A DS agent can be a machine or a human.  Human agents think like humans, but might act like humans or like machines, depending on the data task they have been assigned. 

\noindent
The expertise required for a DS agent depends on:
\begin{enumerate}
\item[(a)] the focus (which of the $5$D complexities are we interested in? Which parts of the data life cycle?), 
\item[(b)] the {task-based} goals (what are the specific {task categorization} ?), and 
\item[(c)] the ${\cal F}$-induced restriction of the data universe. This last point implies that we can define an ${\cal F}$-induced agent, denoted ${\cal F}-{\cal A}$, as an agent with specific skills and objectives
induced by the integration of ${\cal F}$  with the EDS ecosystem. In other words, we define the ${\cal F}$-induced agent, ${\cal F}-{\cal A}$, as an agent whose competencies—such as task decomposition, information retrieval, and interaction within the EDS ecosystem—are determined by the functions and constraints encoded in ${\cal F}$.
\end{enumerate}
We claim that for a given challenge, a number of data agents is chosen from the total number of agents, according to the atomization into goals, which imply tasks and skills. To avoid formalism obfuscation, we claim that every agent can perform a given number of actions. For a given challenge and related atomization, interactions between agents are also considered as actions. Actions are coordinated (by one or more agents) to  ensure that they are consistent  with respect to challenge and mission.

\subsection{Data Scientist}
The definition of data agents, in concert with their actions, {is open to} the following formalization that is central to this paper. 
\begin{postulate} \label{post-data-scientist}
For a given mission and challenge, a data scientist is an entity composed by 
\begin{enumerate}
\item the agent(s) that perform(s) {challenges} atomization into {tasks and goals};
\item a given number of {DS} agents, each with their actions, tasks, and goals, and with their interactions;
\item the agent(s) that coordinate(s) the data agents, to  {ensure} that the delivered results are consistent with the challenge and mission. 
\end{enumerate}
\end{postulate}
\begin{figure}
\scalebox{.8}{
\begin{tikzpicture}[thick, every node/.style={scale=0.8},
    node distance=1.5cm and 2cm,
    box/.style={rectangle, draw, rounded corners, minimum width=2.5cm, minimum height=1cm, text centered},
    arrow/.style={-{Latex[length=2mm]}},
    bidirarrow/.style={{Latex[length=3mm]}-{Latex[length=3mm]}},
]

\node (who) [box] {{\sc DATA UNIVERSE}};
\node (mission) [box, below=of who] {{\sc Mission/Vision}};
\node (challenges) [box, below=of mission] {{\sc Challenges} };
\node (goals) [box, below=of challenges] {{\sc Tasks/Goals/Assignments} };
\node (data_agents) [box, below=of goals] {{\sc Data Agents}};
\node (data_scientist) [box, right=3.5cm of data_agents] {{\sc Data scientist}};
\node (data_universe) [box, right=3.5cm of data_scientist] {{\sc DUD }};
\node (table1) [box, right=9.7cm of who] {Table \ref{table1}};

\draw[arrow] (who) -- node[midway, right] {{\sc COMPANIES $\&$  ACADEMIA}} (mission);
\draw[arrow] (mission) -- node[midway, right] {INSPIRE} (challenges);
\draw[arrow] (challenges) -- node[midway, right] {ATOMIZATION} (goals);
\draw[arrow] (goals) -- node[midway, right] {DEFINE SKILLS}(data_agents);
\draw[arrow] (data_agents) -- node[midway, below] { ORGANIZED AS
} (data_scientist);
\draw[arrow] (data_scientist) -- node[midway, below] {DELIVER RESULTS} (data_universe);
\draw[arrow, dashed] (data_universe) to[bend right=30] node[midway, right] {{\sc UPDATES}} (who);
\draw[bidirarrow,  dashed] (data_scientist) to[bend right=35] node[below] {MISSION CONSISTENCY} (mission);
\draw[bidirarrow, dashed] (who.east) -- (table1.west) node[midway, above] {ORGANIZED AS};
\end{tikzpicture}}
\caption{A scheme of the route from {\em the data universe to DUD via missions}.}
\label{fig_scheme}
\end{figure}
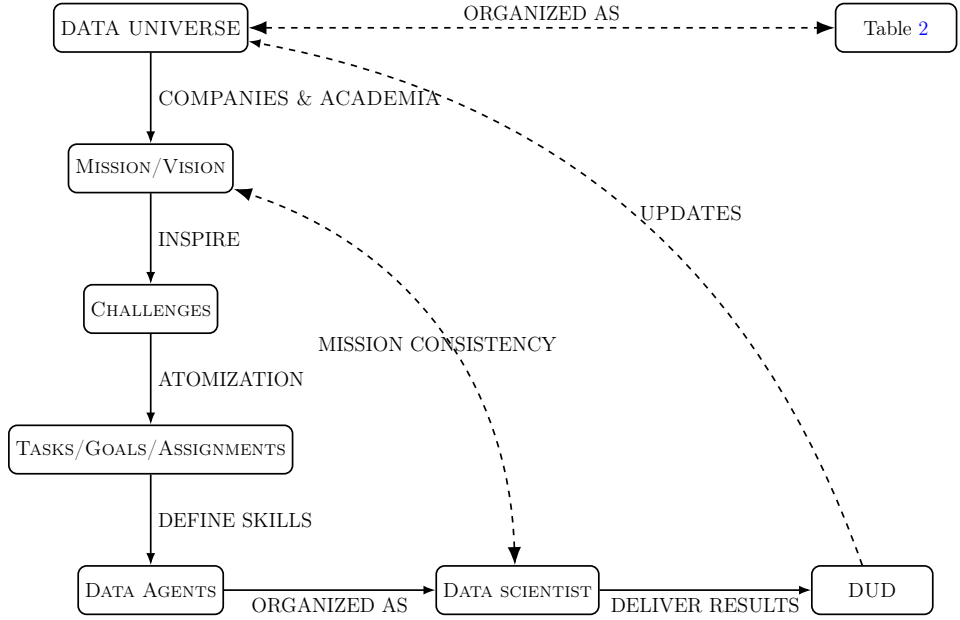
Figure \ref{fig_scheme} depicts the route from the {\em data universe to {\rm DUD} via missions}, where DUD stands for data universe discoveries.
The scheme generator starts with the data universe. Missions and visions might come from corporation, academia, or both. Missions imply a collection of challenges, organized through tasks, and goals (atomization). 
Atomization of the data science tasks is achieved through atomization of the data universe, through what we call the $5$D complexities, in concert with the atomic actions that are specific to the data life cycle. This allows to define specific skills, and to identify the DS agents that are part of the team. This is not enough to determine the corresponding data scientist: the {\em ensembling} requires a nontrivial coordination of the individual tasks performed by the agents. Data scientists deliver results that prompt a data universe update, from the data-discovery analysis, which in turn triggers new challenges, and consequently missions/visions updates.  

\begin{figure}
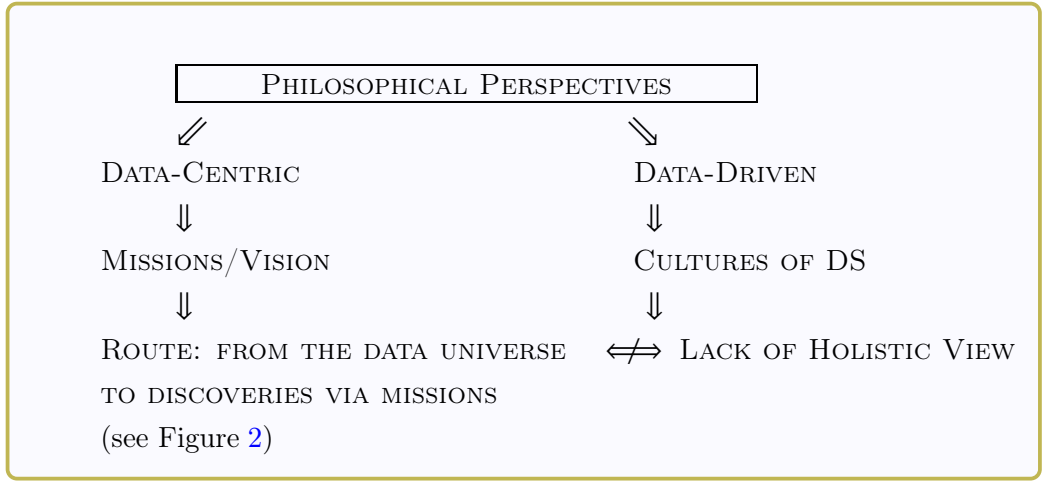
  
\small
\begin{tcolorbox}[colback=blue!2!white,colframe=yellow!70!black]
\begin{eqnarray*}
&& \hspace{1cm}
\text{\fbox{\hspace{1cm}{\sc Philosophical Perspectives}\hspace{1cm}}} \\
&& \hspace{1cm} \Swarrow \hspace{5.5cm}  \Searrow \\
&& \text{{\sc Data-Centric}} \hspace{4.4cm } \text{{\sc Data-Driven}} \\
&& \hspace{1cm} \Downarrow \hspace{6cm} \Downarrow \\
&& \text{{\sc Missions/Vision}}   \hspace{4cm} \text{{\sc Cultures of DS}} \\ 
&& \hspace{1cm} \Downarrow \hspace{6cm} \Downarrow \\
&& \text{{\sc {Route: from the data universe  }
}} \hspace{0cm}\centernot\iff \hspace{0cm} \text{{\sc Lack of Holistic View}}\\
&& \text{{\sc {to discoveries via missions}}}\\
&& \text{(see Figure}  \; \ref{fig_scheme}\text{)}
\end{eqnarray*}
\end{tcolorbox}
\caption{Illustration of two different philosophical perspectives in data science. Right: culture-specific approaches (statistical, algorithmic, domain-first) excel locally but lack a coordinator across 5D×DLC; EDS plays that coordinating role.}\label{fig-Philosophical}
\end{figure}
The scheme in Figure \ref{fig-Philosophical} explains the difference between the data-centric perspectives {\em versus} the data-driven  philosophical ones. The former triggers our route {\em from the data universe to discoveries via missions}, 
as in Figure \ref{fig_scheme}.

\section{Discussion} \label{sec7}

This paper proposes a definition of data science, based on the 5D definition, a five-point data life cycle, and offers a rough classification of the data universe. This allows defining concepts such as EDS, DS Agent and Data Scientist. 

Central to our framework are the concepts of EDS, data agents, and the pan-data science, as well as their interactions from which it stems a modular and adaptive model for integrating heterogeneous sources, methods, and goals. \\
Our key contributions can be resumed as: \\
1. A task-based fusion model that is substantially based on the five dimensions of complexity (5Ds) and a five-phase data life cycle. \\
2. An agent-based formulation where humans and machines form a distributed intelligence system. \\

We argue that the fusions presented in this paper will generate more fusions and new architectures, that will be applicable in a wealth of different domains. These will include AI governance, scientific discovery, and many others, such as sensor networks and auditing systems.

Table~\ref{tab:eds-comparison} is a comparative table, that illustrates the addition of this paper with respect to other definitions of data science.

\begin{table}[tb]
\centering
\scriptsize
\caption{\label{tab:eds-comparison}Positioning of our Essential Data Science (EDS) view against prior framings. }
\begin{adjustbox}{width=.99\textwidth,center}
\begin{tabular}{p{0.17\linewidth} p{0.19\linewidth} p{0.18\linewidth} p{0.20\linewidth} p{0.20\linewidth}}
\toprule
\textbf{Framework} & \textbf{Scope } & \textbf{Unit of analysis} & \textbf{Mission } & \textbf{Lifecycle \& agents } \\
\midrule
\textbf{Breiman (2001) \citep{breiman2001statistical}} &
Statistical vs. algorithmic modeling; prediction performance and accuracy & 
Prediction task on a dataset. & 
Implicit: mission often reduced to out-of-sample error; KPIs mainly predictive scores & 
Lifecycle largely implicit; individual analyst and algorithmic tools, no explicit multi-agent coordination \\
\midrule
\textbf{Donoho (2017) \citep{donoho201750}} & 
Broad DS activities (data gathering, wrangling, exploration, modeling, visualization, inference, communication) &
Practices and competencies & 
Recognizes impact of communication but not mission-driven KPIs as central objects & 
End-to-end orientation; roles are human-centric; orchestration is descriptive rather than formalized \\
\midrule
\textbf{Gray (2009) \citep{gray2000rules}} &
Data-intensive scientific discovery and e-Science infrastructure &
Scientific programs and data repositories & 
Discovery oriented goals; KPIs tied to scientific outcomes and sharing & 
Lifecycle stresses curation and cyberinfrastructure; orchestration via tools/platforms, not task-level agents \\
\midrule
\textbf{van der Aalst \citep{aalst2016process}} &
Event-log analysis for conformance, performance, enhancement of processes &
Event logs, process models, conformance metrics &
Mission/KPIs framed as process conformance and performance &
Lifecycle centered on process mining pipeline; orchestration by algorithms + analysts within that pipeline \\
\midrule
\textbf{Ours: EDS} &
Mission to challenge to task atomization via \emph{5D} (domain, structure, cardinality, causality, ethics) and \emph{DLC} (design, capture, quality, analysis, communication) &
5D$\times$DLC pairings and the induced task graph &
\textbf{Explicit and primary}: mission-defined KPIs (proper scores, coverage, equity) drive design, analysis, and decisions &
\textbf{Decision-anchored lifecycle} with \emph{DS agents} (human + AI) assigned to tasks; accountability, calibration loops, and cross-agent coordination made explicit \\
\bottomrule
\end{tabular}
\end{adjustbox}
\end{table}

We conclude with a list of takeaways, as per the following ten points. 
\begin{enumerate}
\item data science that do not rely on other disciplines will bring many benefits. 
{Separating data science from AI is both conceptually and practically valuable. While AI increasingly influences all areas of science and technology -- particularly through the deployment of AI agents and the utilization of advanced computational resources such as foundation models and large language models --, we argue that establishing robust, independent foundations for data science offers significant advantages. By focusing on data science as a discipline in its own right, free from reliance on other fields, we can unlock its full potential and foster greater innovation.\\}
The interactions between data science and AI are certainly a target for future papers.
\item  Empirical machine learning is a pillar of computational EDS, but not of all EDS. This fact should mitigate the apparent confusion in the literature, where data science has been increasingly identified with machine learning. 
\item Big data cover a fundamental importance in today's life. However, data science is concerned with all the complexities of the data universe, big (or small) data being only one of them. 
\item Data science requires holistic views triggering real missions, inspiring  the mission-to-mission route, {via the route {\em data universe to discoveries via missions.}} 

\item We hope that some smart information theorist will read this paper and provide an adaptation of the Bengio-Malkin routine for the case of useful DUDs. 
\item After reading this paper, do you still think that you are a data scientist? We understand that 50 years of confusion created a {\em garden of wishful thinking} at both academic and corporate level. We hope that at least some work will be devoted to explore the consequences of this paper. 
\item If you accept our theory of DS agents, what kind of data agent are you?
\item If you {\bf did not} accept our theory of DS agents, then what kind of data agent are you?
\item Which kind of data scientist would you like to belong to?
\item Do not torture your data until they confess. Be gentle, our theory (and data ethics) gives data a proper essence and dignity.
\end{enumerate}
Our approach will hopefully increase the collaboration between the two planets and communities. A natural step is to provide a detailed analysis between DS and AI agents, with the purpose of marking the distinctive features while highlighting the potential synergies.

{
\begin{acks}
The authors would like to thank Professors Dimitrios Kyristsis, Leonardo Palmisano, and Jean Ponce for their careful reading of our paper and for their invaluable feedback. The authors are grateful to the Editors and to a Referee whose detailed comments allowed for a considerably improved version of the manuscript. 
\end{acks}}

{
\subsection*{Author contributions}
EP, REM,  and LN conceptualized the study, developed the theoretical framework, and wrote the majority of the manuscript. FH and HS reviewed the manuscript, provided critical feedback, and offered valuable comments that improved the final version. All authors read and approved the final manuscript.
}

{\subsection*{Funding statement}
This research did not receive any specific grant from funding agencies in the public, commercial, or not-for-profit sectors.
}

\theendnotes

\bibliographystyle{unsrtnat}
\bibliography{references.bib}

\begin{thebibliography}{31}
\providecommand{\natexlab}[1]{#1}
\providecommand{\url}[1]{\texttt{#1}}
\expandafter\ifx\csname urlstyle\endcsname\relax
  \providecommand{\doi}[1]{doi: #1}\else
  \providecommand{\doi}{doi: \begingroup \urlstyle{rm}\Url}\fi

\bibitem[Breiman(2001)]{breiman2001statistical}
Leo Breiman.
\newblock Statistical modeling: The two cultures (with comments and a rejoinder
  by the author).
\newblock \emph{Statistical science}, 16\penalty0 (3):\penalty0 199--231, 2001.

\bibitem[Liberman and F{\"o}rster(2009)]{liberman2009effect}
Nira Liberman and Jens F{\"o}rster.
\newblock The effect of psychological distance on perceptual level of
  construal.
\newblock \emph{Cognitive science}, 33\penalty0 (7):\penalty0 1330--1341, 2009.

\bibitem[Donoho(2019)]{LectureA}
David Donoho.
\newblock Deepnet spectra and the two cultures of data science.
\newblock Al Kindi Distinguished Lecture, 2019.

\bibitem[Hern{\'a}n et~al.(2019)Hern{\'a}n, Hsu, and Healy]{hernan2019chance}
{Miguel A.} Hern{\'a}n, John Hsu, and Brian Healy.
\newblock A second chance to get causal inference right: a classification of
  data science tasks.
\newblock \emph{CHANCE}, 32\penalty0 (1):\penalty0 42--49, 2019.

\bibitem[Hey et~al.(2009)Hey, Tansley, Tolle, et~al.]{hey2009fourth}
Tony Hey, Stewart Tansley, Kristin~Michele Tolle, et~al.
\newblock \emph{The fourth paradigm: data-intensive scientific discovery},
  volume~1.
\newblock Microsoft research Redmond, WA, 2009.

\bibitem[Donoho(2024)]{donoho2024data}
David Donoho.
\newblock Data science at the singularity.
\newblock \emph{Harvard Data Science Review}, 6\penalty0 (1), 2024.

\bibitem[Aalst(2016)]{aalst2016process}
Wil van~der Aalst.
\newblock Process mining: data science in action.
\newblock \emph{(No Title)}, 2016.

\bibitem[Donoho(2017)]{donoho201750}
David Donoho.
\newblock 50 years of data science.
\newblock \emph{Journal of Computational and Graphical Statistics}, 26\penalty0
  (4):\penalty0 745--766, 2017.

\bibitem[Tukey(1962)]{tukey1962future}
John~W Tukey.
\newblock The future of data analysis.
\newblock In \emph{Breakthroughs in Statistics: Methodology and Distribution},
  pages 408--452. Springer, 1962.

\bibitem[Cleveland(2001)]{cleveland2001data}
William~S Cleveland.
\newblock Data science: an action plan for expanding the technical areas of the
  field of statistics.
\newblock \emph{International statistical review}, 69\penalty0 (1):\penalty0
  21--26, 2001.

\bibitem[Meng(2019)]{meng2019data}
Xiao-Li Meng.
\newblock Data science: An artificial ecosystem.
\newblock \emph{Harvard Data Science Review}, 1\penalty0 (1):\penalty0
  10--1162, 2019.

\bibitem[Jordan(2019)]{jordan2019artificial}
Michael~I Jordan.
\newblock Artificial intelligence—the revolution hasn’t happened yet.
\newblock \emph{Harvard Data Science Review}, 1\penalty0 (1):\penalty0 1--9,
  2019.

\bibitem[Aamodt and Nyg{\aa}rd(1995)]{aamodt1995different}
Agnar Aamodt and Mads Nyg{\aa}rd.
\newblock Different roles and mutual dependencies of data, information, and
  knowledge—an {AI} perspective on their integration.
\newblock \emph{Data \& Knowledge Engineering}, 16\penalty0 (3):\penalty0
  191--222, 1995.

\bibitem[Leonelli(2019)]{leonelli2019data}
Sabina Leonelli.
\newblock Data {Governance} is {Key} to {Interpretation}: Reconceptualizing
  {Data} in {Data} {Science}.
\newblock \emph{Harvard Data Science Review}, 1\penalty0 (1), 7 2019.
\newblock https://hdsr.mitpress.mit.edu/pub/4ovhpe3v.

\bibitem[Berman et~al.(2018)Berman, Rutenbar, Hailpern, Christensen, Davidson,
  Estrin, Franklin, Martonosi, Raghavan, Stodden, et~al.]{berman2018realizing}
Francine Berman, Rob Rutenbar, Brent Hailpern, Henrik Christensen, Susan
  Davidson, Deborah Estrin, Michael Franklin, Margaret Martonosi, Padma
  Raghavan, Victoria Stodden, et~al.
\newblock Realizing the potential of data science.
\newblock \emph{Communications of the ACM}, 61\penalty0 (4):\penalty0 67--72,
  2018.

\bibitem[Pearl(2009)]{pearl2009causality}
Judea Pearl.
\newblock \emph{Causality}.
\newblock Cambridge university press, 2009.

\bibitem[Hernán and Robins(2020)]{Hernan-Robins-2020}
MA~Hernán and JM~Robins.
\newblock \emph{Causal Inference: What If}.
\newblock Boca Raton: Chapman \& Hall/CRC., 2020.

\bibitem[L{\'o}pez~de Prado(2023)]{de2023causal}
Marcos~M L{\'o}pez~de Prado.
\newblock \emph{Causal Factor Investing: Can Factor Investing Become
  Scientific?}
\newblock Cambridge University Press, 2023.

\bibitem[Saltz and Dewar(2019)]{saltz2019data}
Jeffrey~S Saltz and Neil Dewar.
\newblock Data science ethical considerations: a systematic literature review
  and proposed project framework.
\newblock \emph{Ethics and Information Technology}, 21:\penalty0 197--208,
  2019.

\bibitem[Barocas and Boyd(2017)]{barocas2017engaging}
Solon Barocas and Danah Boyd.
\newblock Engaging the ethics of data science in practice.
\newblock \emph{Communications of the ACM}, 60\penalty0 (11):\penalty0 23--25,
  2017.

\bibitem[Okorie et~al.(2024)Okorie, Udeh, Adaga, DaraOjimba, and
  Oriekhoe]{okorie2024ethical}
Gold~Nmesoma Okorie, Chioma~Ann Udeh, Ejuma~Martha Adaga, Obinna~Donald
  DaraOjimba, and Osato~Itohan Oriekhoe.
\newblock Ethical considerations in data collection and analysis: a review:
  investigating ethical practices and challenges in modern data collection and
  analysis.
\newblock \emph{International Journal of Applied Research in Social Sciences},
  6\penalty0 (1):\penalty0 1--22, 2024.

\bibitem[Floridi and Taddeo(2016)]{floridi2016data}
Luciano Floridi and Mariarosaria Taddeo.
\newblock What is data ethics?
\newblock \emph{Philosophical Transactions of the Royal Society A:
  Mathematical, Physical and Engineering Sciences}, 374\penalty0
  (2083):\penalty0 20160360, 2016.

\bibitem[Wang et~al.(2024)Wang, Liu, Li, Lin, Sindakis, and
  Aggarwal]{wang2024overview}
Jingran Wang, Yi~Liu, Peigong Li, Zhenxing Lin, Stavros Sindakis, and Sakshi
  Aggarwal.
\newblock Overview of data quality: Examining the dimensions, antecedents, and
  impacts of data quality.
\newblock \emph{Journal of the Knowledge Economy}, 15\penalty0 (1):\penalty0
  1159--1178, 2024.

\bibitem[Garc{\'\i}a et~al.(2015)Garc{\'\i}a, Luengo, and
  Herrera]{garcia2015data}
Salvador Garc{\'\i}a, Juli{\'a}n Luengo, and Francisco Herrera.
\newblock \emph{Data preprocessing in data mining}, volume~72.
\newblock Springer, 2015.

\bibitem[Borgman(2019)]{borgman2019lives}
Christine~L Borgman.
\newblock The lives and after lives of data.
\newblock \emph{Harvard Data Science Review}, 1\penalty0 (1):\penalty0
  10--1162, 2019.

\bibitem[Wing(2019)]{wing2019data}
Jeannette~M Wing.
\newblock The data life cycle.
\newblock \emph{Harvard Data Science Review}, 1\penalty0 (1):\penalty0 6, 2019.

\bibitem[Borgman(2017)]{borgman2017big}
Christine~L Borgman.
\newblock \emph{Big data, little data, no data: Scholarship in the networked
  world}.
\newblock MIT press, 2017.

\bibitem[Murphy(1998)]{murphy1998concepts}
JJ~Murphy.
\newblock The concepts of vision and mission revisited.
\newblock \emph{Southern African Business Review}, 2\penalty0 (1):\penalty0
  27--33, 1998.

\bibitem[Bennis and Nanus(1985)]{bennis1985strategies}
Warren Bennis and Burt Nanus.
\newblock The strategies for taking charge.
\newblock \emph{Leaders, New York: Harper. Row}, 41, 1985.

\bibitem[Quigley(1994)]{quigley1994vision}
Joseph~V Quigley.
\newblock Vision: How leaders develop it, share it, and sustain it.
\newblock \emph{Business Horizons}, 37\penalty0 (5):\penalty0 37--42, 1994.

\bibitem[Gray and Shenoy(2000)]{gray2000rules}
Jim Gray and Prashant Shenoy.
\newblock Rules of thumb in data engineering.
\newblock In \emph{Proceedings of 16th international conference on data
  engineering (Cat. No. 00CB37073)}, pages 3--10. IEEE, 2000.

\end{thebibliography}

\end{document}